\documentclass{article}

\usepackage{arxiv}

\usepackage[utf8]{inputenc} % allow utf-8 input
\usepackage[T1]{fontenc}    % use 8-bit T1 fonts
\usepackage{hyperref}       % hyperlinks
\usepackage{url}            % simple URL typesetting
\usepackage{booktabs}       % professional-quality tables
\usepackage{amsfonts}       % blackboard math symbols
\usepackage{nicefrac}       % compact symbols for 1/2, etc.
\usepackage{microtype}      % microtypography
\usepackage{lipsum}		% Can be removed after putting your text content
\usepackage{graphicx}
\usepackage{doi}
\usepackage{graphicx}
\graphicspath{ {./} }

\title{Neural Architecture Search for Intel Movidius VPU}

\author{ \hspace{1mm}Qian Xu \\
	Intel \\
	\texttt{xu.qian@intel.com} \\
	\And
	\hspace{1mm}Victor Li \\
	Intel \\
	\texttt{victor.y.li@intel.com} \\
    \And
    \hspace{1mm}Crews Darren S \\
	Intel \\
	\texttt{darren.s.crews@intel.com} \\
}

\renewcommand{\shorttitle}{\textit{arXiv} Template}

\begin{document}
\maketitle

\begin{abstract}
Hardware-aware Neural Architecture Search (NAS) technologies have been proposed to automate and speed up model design to meet both quality and inference efficiency requirements on a given hardware.  Prior arts have shown the capability of NAS on hardware specific network design. In this whitepaper, we further extend the use of NAS to Intel Movidius VPU (Vision Processor Units). To determine the hardware-cost to be incorporated into the NAS process, we introduced two methods: pre-collected hardware-cost on device and device-specific hardware-cost model VPUNN. With the help of NAS, for classification task on VPU, we can achieve 1.3x fps acceleration over Mobilenet-v2-1.4 and 2.2x acceleration over Resnet50 with the same accuracy score. For super resolution task on VPU, we can achieve 1.08x PSNR and 6x higher fps compared with EDSR3.
\end{abstract}

% keywords can be removed

\section{Introduction}
Intel Movidius VPU enable demanding computer vision and AI workloads with efficiency. By coupling highly parallel programmable compute with workload-specific AI hardware acceleration in a unique architecture that minimizes data movement, Movidius VPUs achieve a balance of power efficiency, and compute performance.

\begin{figure}[h!]
    \centering
    \includegraphics[scale=0.35]{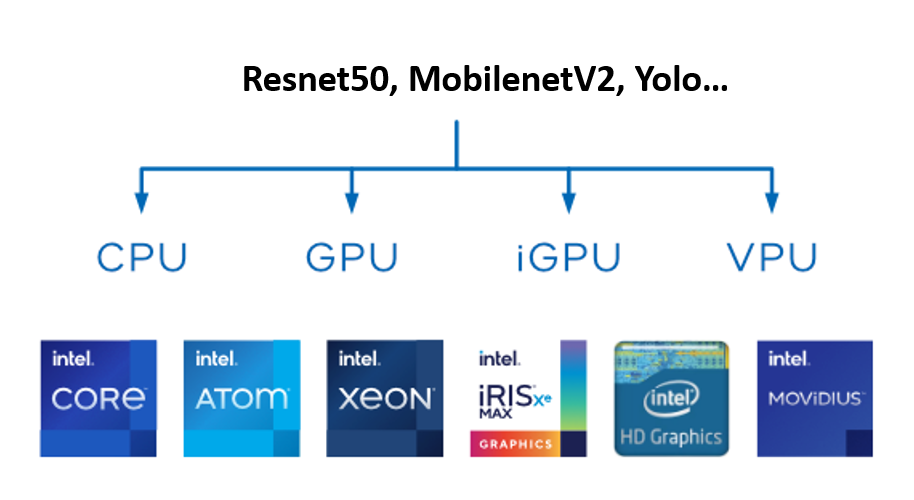}
    \hspace{0in}
    \includegraphics[scale=0.37]{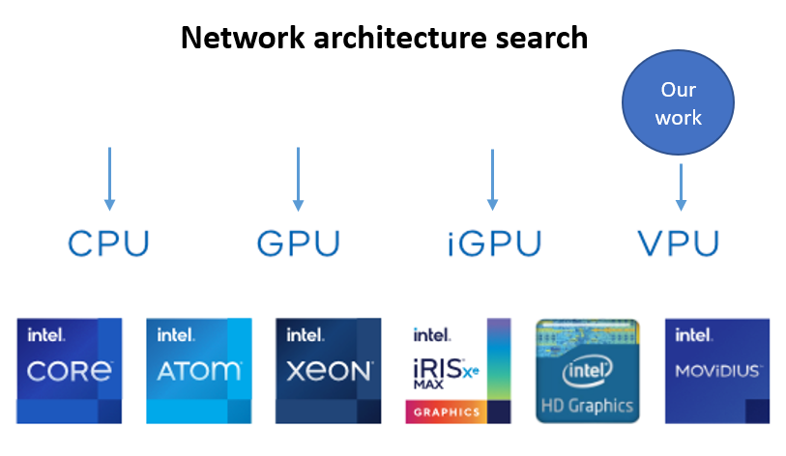}
    \caption{Neural Architecture Search for Hardware devices}
    \label{fig.1}
\end{figure}

But the AI models from customers are usually generally built and not designed for a specific hardware as fig.\ref{fig.1} left shows. Due to the different designs of various AI accelerators, general models can’t fully utilize hardware’s capability.  That gives the chance to design better models for hardware: higher fps at same accuracy level or higher accuracy at same fps.  However, even for hardware specialists, the design space of possible networks is still extremely large and impossible for handcrafting. \cite{https://doi.org/10.48550/arxiv.2101.09336} \cite{https://doi.org/10.48550/arxiv.1910.11609}

In this whitepaper, we address the huge design space by incorporating the Neural Architecture Search methods as Fig.\ref{fig.1} right shows.

Specially, to optimize directly for the target hardware VPU, we need to include the latency of the model on the hardware in the loss function. In this whitepaper, we propose two methods to get the latency of the network on Hardware: pre-collected hardware-cost on device \cite{https://doi.org/10.48550/arxiv.2103.10584} and device-specific hardware-cost model VPUNN \cite{https://doi.org/10.48550/arxiv.2205.04586}. 

In our experiments with image classfication tasks, we can achieve 1.3x fps acceleration over Mobilenet-v2-1.4 \cite{https://doi.org/10.48550/arxiv.1704.04861} and 2.2x acceleration over Resnet50 \cite{https://doi.org/10.48550/arxiv.1512.03385} with the same accuracy score. For super resolution, we can achieve 1.08x PSNR and higher fps compared with EDSR \cite{https://doi.org/10.48550/arxiv.1707.02921}. Our contributions can be summarized as follows:

\begin{itemize}
\item	Use ProxylessNAS \cite{https://doi.org/10.48550/arxiv.1812.00332} for image classification and super resolution tasks and generate ready-to-use models for customers of Intel VPU. The models beat original workloads in both speed and accuracy.

\item	We present two methods for model profiling on hardware: pre-collected hardware-cost on device and device-specific hardware-cost model VPUNN. Pre-collected hardware-cost look-up is more accurate but limited to VPU experts. VPUNN is open-sourced profiling tool of VPU and can work out-of-the-box for customers that don’t have direct access to VPU hardware.

\item	From the searched networks, we have some intuitive guidelines for designing networks for Intel VPU. Customers of Intel VPU can follow these guidelines when optimizing networks for VPU.
\end{itemize}

\section{Related works}
\label{sec:headings}

\subsection{ProxylessNAS}

\begin{figure}[h!]
    \centering
    \includegraphics[scale=0.4]{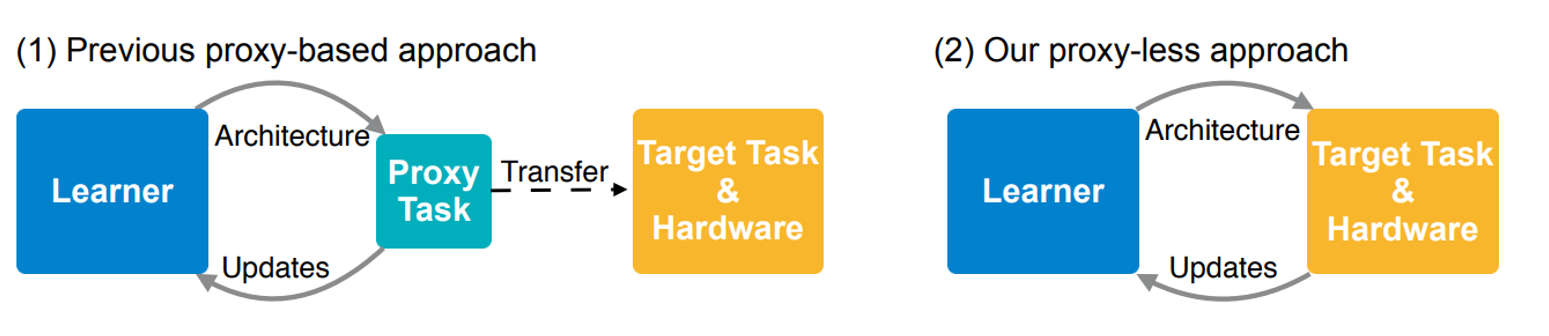}
    \caption{illustration of ProxylessNAS}
    \label{fig.2}
\end{figure}

ProxylessNAS \cite{https://doi.org/10.48550/arxiv.1812.00332} is the first algorithm that directly learns architectures on the largescale dataset (e.g. ImageNet) without any proxy while still allowing a large candidate set and removing the restriction of repeating blocks as fig\ref{fig.2} shows. It effectively enlarged the search space and achieved better performance. It achieved state-of-the-art accuracy performances on CIFAR-10 and ImageNet under latency constraints on different hardware platforms (GPU, CPU and mobile phone). In this whitepaper, we are going to extend the use of ProxylessNas to Intel VPU.

\subsection{Latency Profiling}

Measuring the latency on-device is accurate but not ideal for scalable neural architecture search because this is slow and expensive.

To determine the hardware-cost to be incorporated into the NAS process, existing works mostly adopt either pre-collected hardware-cost on device or device-specific hardware-cost models. The former can be time-consuming due to the required knowledge of the device’s compilation method and how to set up the measurement pipeline. But the former one is more accurate since it is measured directly on actual hardware. For the second method, there is already a cost model for performance projection of VPU named VPUNN. This is a neural network-based cost model trained on low-level task profiling that consistently outperforms the state-of-the-art cost modeling in Intel’s line of VPU processors. The features to the model include (i) type of the operator (ii) input and output feature map size (iii) other attributes like kernel size, stride for convolution and so on.

\section{Solution}

Neural architecture search (NAS) focuses on automating the architecture design process.  Early NAS methods search for high-accuracy architectures without taking hardware efficiency into consideration. Therefore, the produced architectures are not efficient for inference.  Recent hardware-aware NAS methods directly incorporate the hardware feedback into architecture search. In this whitepaper, we utilize ProxylessNAS which takes latency feedback in the training process and optimize directly for latency on target device. To determine the hardware-cost to be incorporated into the NAS process, we introduced two methods: pre-collected hardware-cost on device and device-specific hardware-cost model VPUNN.

\subsection{Path level Pruning and Binarization}

\begin{figure}[h!]
    \centering
    \includegraphics[scale=0.37]{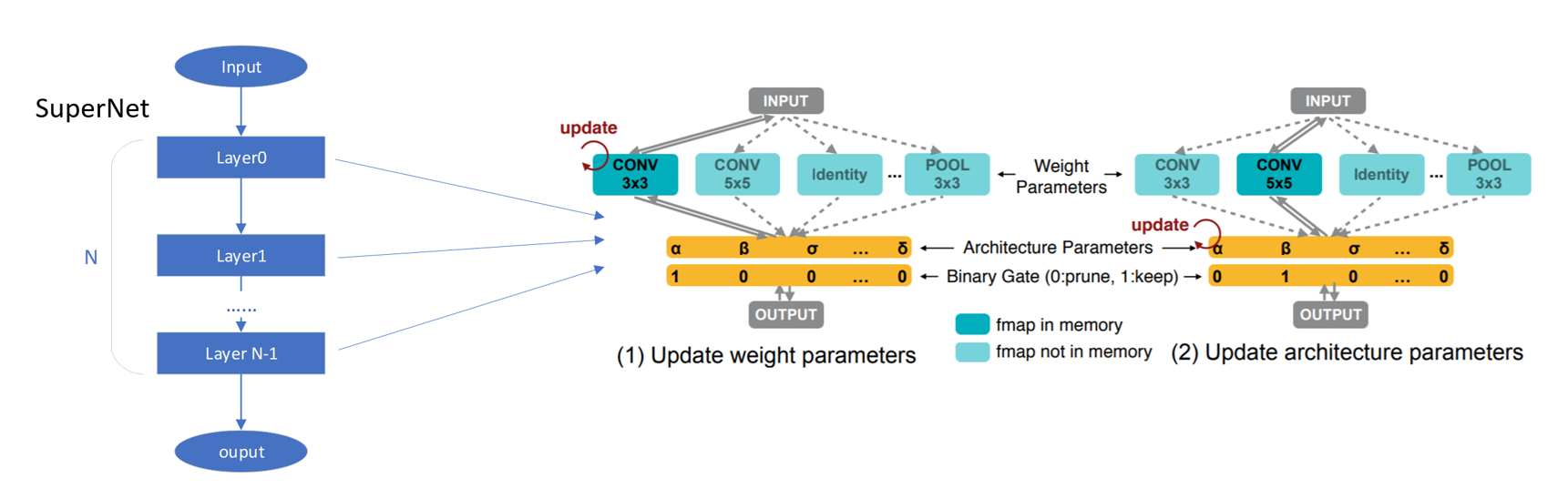}
    \caption{Learning both weight parameters and binarized architecture parameters.}
    \label{fig.3}
\end{figure}

Adopt similar methods from ProxylessNAS, we first construct a over-parameterized network. The super-net consists of multiple stages, as shown in fig.\ref{fig.3}. The super-net is split into N stages. For each stage, we have multiple candidate operators you want to select from. In the case of the figure, we have candidate operators including Conv\_3x3, Conv\_5x5, Identity Op and Pool\_3x3. Then we have a super-net with N stages and for each stage, we have M operators.

The output feature maps of all N stages and M operators are all calculated and stored in the memory, while training a compact model only involves one path. The training time would increase dramatically compared with compact model especially when we have multiple operators in the candidate operator pool.

To reduce memory footprint, ProxylessNAS keep only one path when training the over-parameterized network.

Fig.\ref{fig.3} illustrates the training procedure of the weight parameters and binarized architecture parameters in the over-parameterized network. When training weight parameters, ProxylessNAS first freeze the architecture parameters and stochastically sample binary gates for each batch of input data. Then the weight parameters of active paths are updated via standard gradient descent on the training set (fig.\ref{fig.3} left). When training architecture parameters, the weight parameters are frozen, then we reset the binary gates and update the architecture parameters on the validation set (fig.\ref{fig.3} right). These two update steps are performed in an alternative manner. Once the training of architecture parameters is finished, we can then derive the compact architecture by pruning redundant paths. In this work, we simply choose the path with the highest path weight. 

Unlike weight parameters, the architecture parameters are not directly involved in the computation graph and thereby cannot be updated using the standard gradient descent. So ProxylessNAS introduce a gradient-based approach to learn the architecture parameters. And the gradient w.r.t. architecture parameters can be approximately estimated using:

\begin{equation}
\frac{\partial L}{\partial \alpha_{i}}
=\sum_{j=1}^{N}\frac{\partial L}{\partial p_{j}}\frac{\partial p_{j}}{\partial \alpha_{j}}
=\sum_{j=1}^{N}\frac{\partial L}{\partial g_{j}}\frac{\partial p_{j}}{\partial \alpha_{j}}
\approx\sum_{j=1}^{N}\frac{\partial L}{\partial g_{j}}(\delta_{ij}-p_{i})
\label{eq.1}
\end{equation}

where $ \delta_{ij} = 1$ if $i = j$, and $ \delta_{ij} = 0$ if $i \neq j$. We can update the architecture parameters $\alpha_{i}$ through back-propagation using the upper equation.

\subsection{Making Latency Differentiable}

Besides accuracy, latency is another very important objective when designing efficient neural network architectures for hardware. Unfortunately, unlike accuracy that can be optimized using the gradient of the loss function, latency is non-differentiable. 

\begin{figure}[h!]
    \centering
    \includegraphics[scale=0.35]{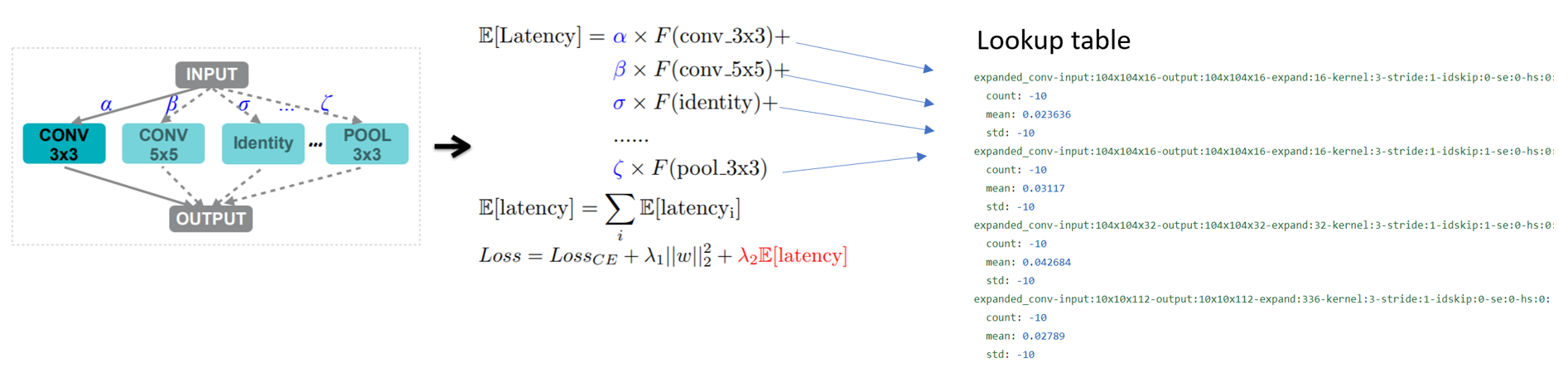}
    \caption{Making latency differentiable by introducing latency regularization loss}
    \label{fig.4}
\end{figure}

As fig.\ref{fig.4} shows, we can estimate the latency of a mixed operator containing multiple candidate operators using:

\begin{equation}
E[latency_{i}]=\sum_{j}^{}p_{i}^{j}\times F(o_{i}^{j})    
\end{equation}
where $E[latency_{i}]$ is the expected latency of the $i^{th}$ block, $p$ is the possibility of selecting a candidate operator, $F()$ denotes the look up table in fig.\ref{fig.4}. Through looking up the latency table through $F({o_{i}^{j}})$, we can get the latency estimation of that candidate operator ${o_{i}^{j}}$ on VPU.

To estimate the latency of the sampled models from the super-net, we model the latency as a continuous function of the neural network dimensions. That means the latency of the sampled network can be calculated by summing up all the selected components. This works well when the scheduling of the target device is linear and the operators are executed sequentially during inference. Then the expected latency of the network is estimated by:

\begin{equation}
E[latency]=\sum_{i}^{}E[latency_{i}])    
\end{equation}

The expected latency is incorporated into the loss function as a regularization loss as show in the following equation:
\begin{equation}
Loss=Loss_{CE} + \lambda_{1}\left\| w \right\|_{2}^{2}+\lambda_{2}E[lantecy]]
\end{equation}

\subsection{Latency Profiling}

As discussed in fig.\ref{fig.4}, the latency of different operators $F({o_{i}^{j}})$ is fetched though a look up table (LUT). In this section, we are going to talk about the two methods we use to get the LUT for VPU.

\begin{figure}[h!]
    \centering
    \includegraphics[scale=0.6]{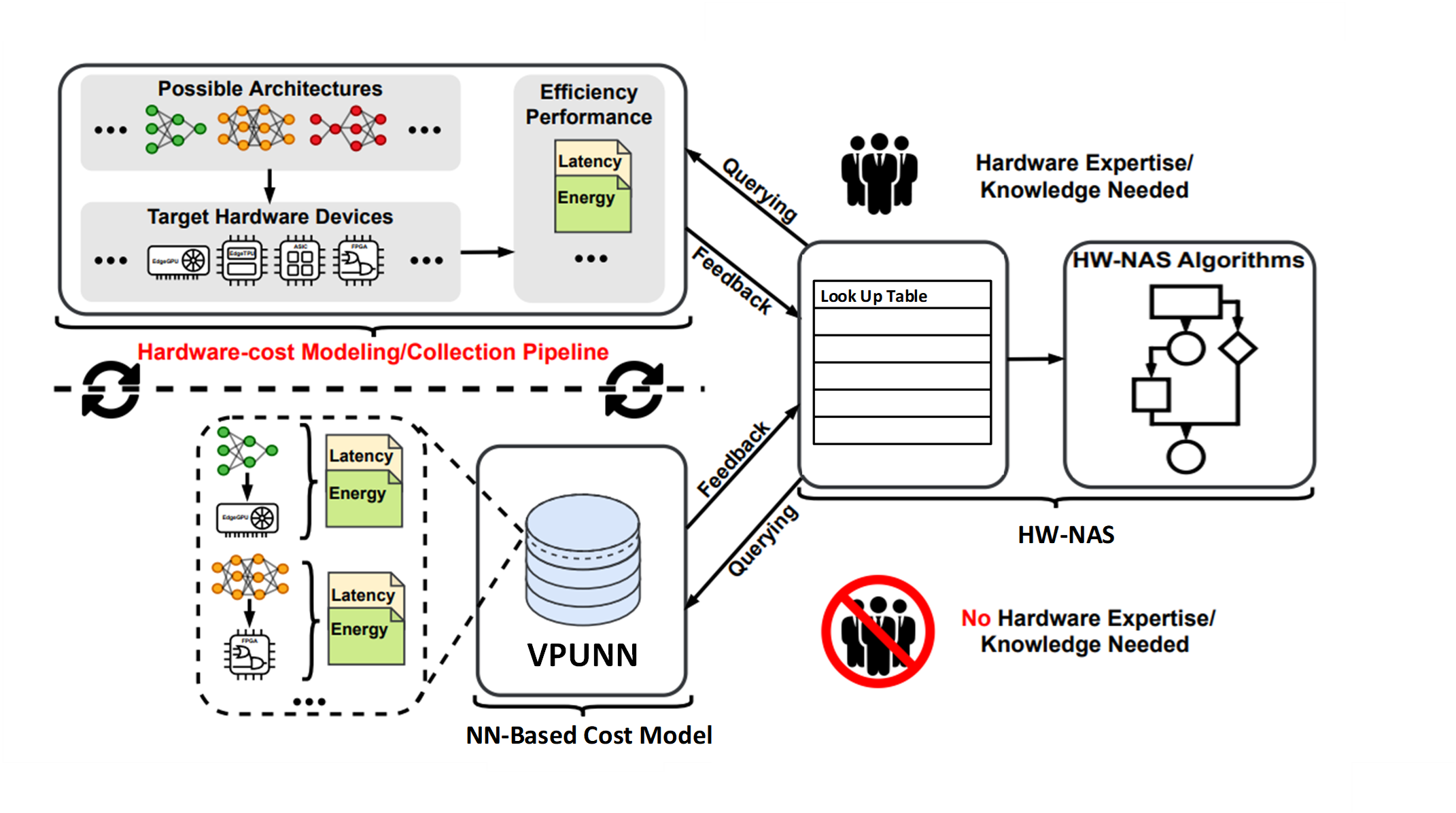}
    \caption{Two methods for latency profile on Hardware}
    \label{fig.5}
\end{figure}

As shown in fig.\ref{fig.5}, we mainly have two methods to project the latency of a operator on hardware. One is by running inference of the operator on actual hardware. Another is by NN-based cost model to estimate the latency.

\subsubsection{Hardware cost collection pipeline}

In this section, we first discuss the pipeline of collecting cost on the hardware.

The most straightforward way to benchmark a layer latency on hardware is by constructing a subgraph with only that operator and run that graph on hardware. But measuring latency of the graph with only one operator has high variance since there are some extra overhead that would has impact on the start time.

\begin{figure}[h!]
    \centering
    \includegraphics[scale=0.2]{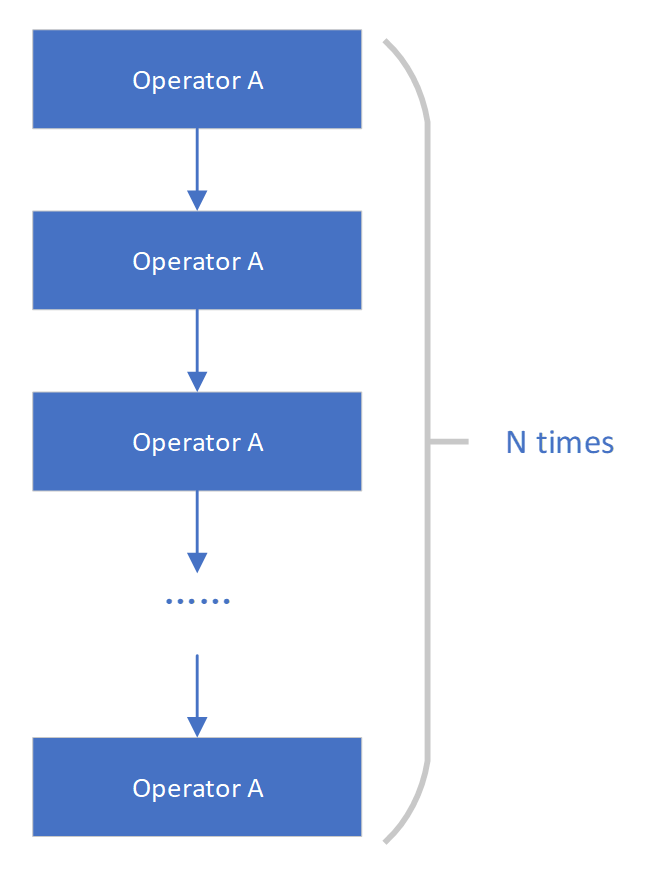}
    \hspace{0in}
    \includegraphics[scale=0.2]{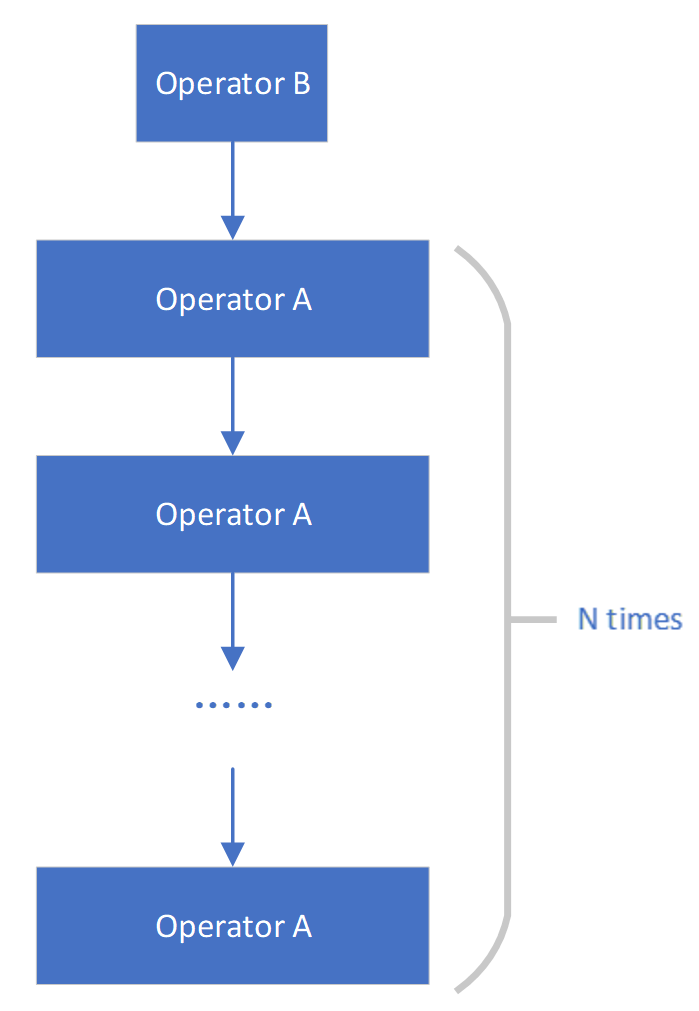}
    \caption{Duplicate the operators to reduce latency variance}
    \label{fig.6}
\end{figure}

To address this issue, we propose to stack the operators to minimize the variances. As fig.\ref{fig.6} left shows, for operators that have the same input and output feature map size, we can directly stack the operators since the operators without any issue. Then we can get the latency of the operator by:

\begin{equation}
F(Op_A) = Latency(Subgraph) / N
\end{equation}

When the operators' input and output feature don't match. we would stack that operator with several other opertors with same input output feature map as fig.\ref{fig.6} right shows. Then the latency is estimated by:

\begin{equation}
F(Op_B) = Latency(Subgraph) - N * F(Op_A)
\end{equation}

\begin{figure}[h!]
    \centering
    \includegraphics[scale=0.3]{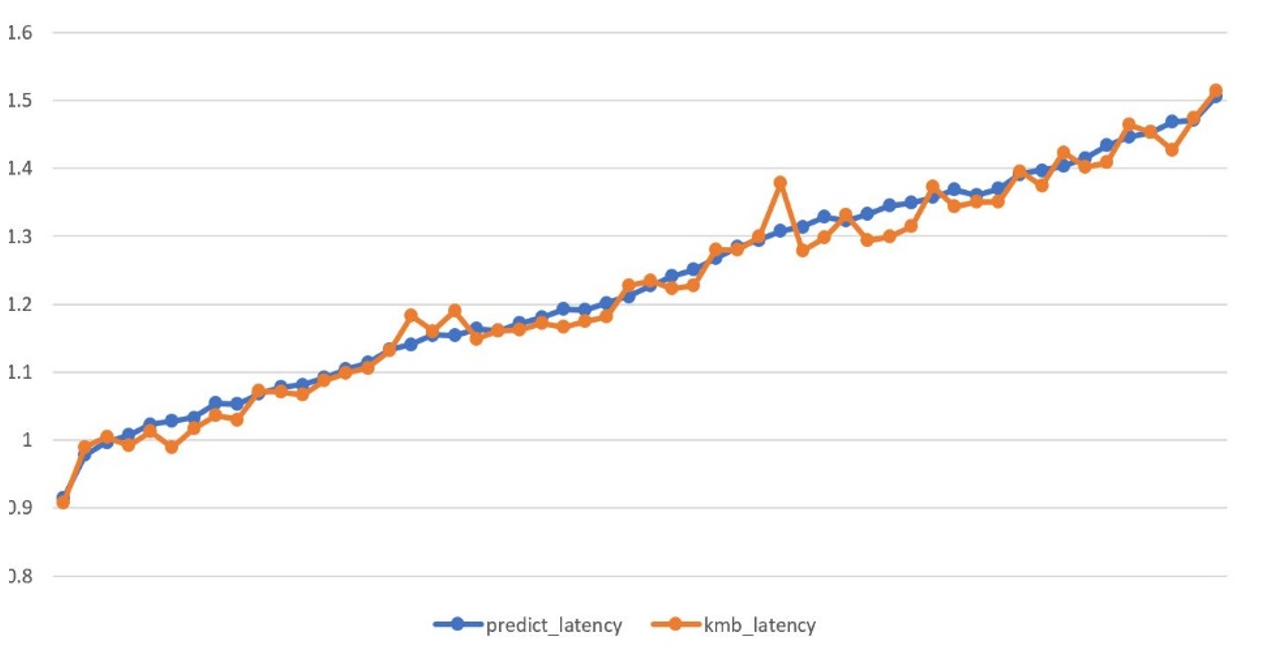}
    \caption{Predicted latency using hardware cost collection pipeline vs Actual latency on VPU}
    \label{fig.7}
\end{figure}

To verify the accuracy of the upper method, we randomly sampled networks from the search space of ImageNet classification super-net. As shown in fig.\ref{fig.7}, we plotted the profiled latency using the upper method and actual network latency tested on VPU. From the figure, the predicted latency basically aligned with the actual latency on the second generation of Movidius VPU KeemBay (KMB).

\subsubsection{NN-Based Cost Model VPUNN}
\begin{figure}[h!]
    \centering
    \includegraphics[scale=0.4]{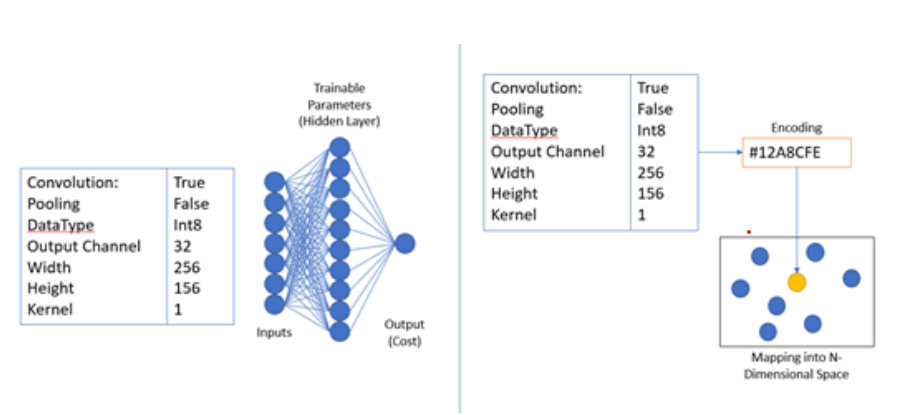}
    \caption{VPUNN: MLP based model to project workload latency}
    \label{fig.8}
\end{figure}

Another way of profiling the layer latency is by device-specific hardware-cost model. There is already an open-sourced model for workload cycles projection of VPU named VPUNN. A simplified diagram can be seen in fig.\ref{fig.8}, VPUNN is basically a two layer fully connected neural network.

This neural network has an input for each relevant aspect of a hardware task’s description, has trained its parameters based on the database entries, and predicts cycle cost as an output. As shown in fig.\ref{fig.8}, VPUNN can predict latency for varies workloads including Convolution, Pooling and so on.

Comparison of the Absolute Percentage Error between previous modelling tools and VPUNN when trained on hardware recordings — versus the actual cycle count of hardware recordings. These results shows the accuracy of VPUNN compared with actual hardware cycles.

\begin{figure}[h!]
    \centering
    \includegraphics[scale=0.28]{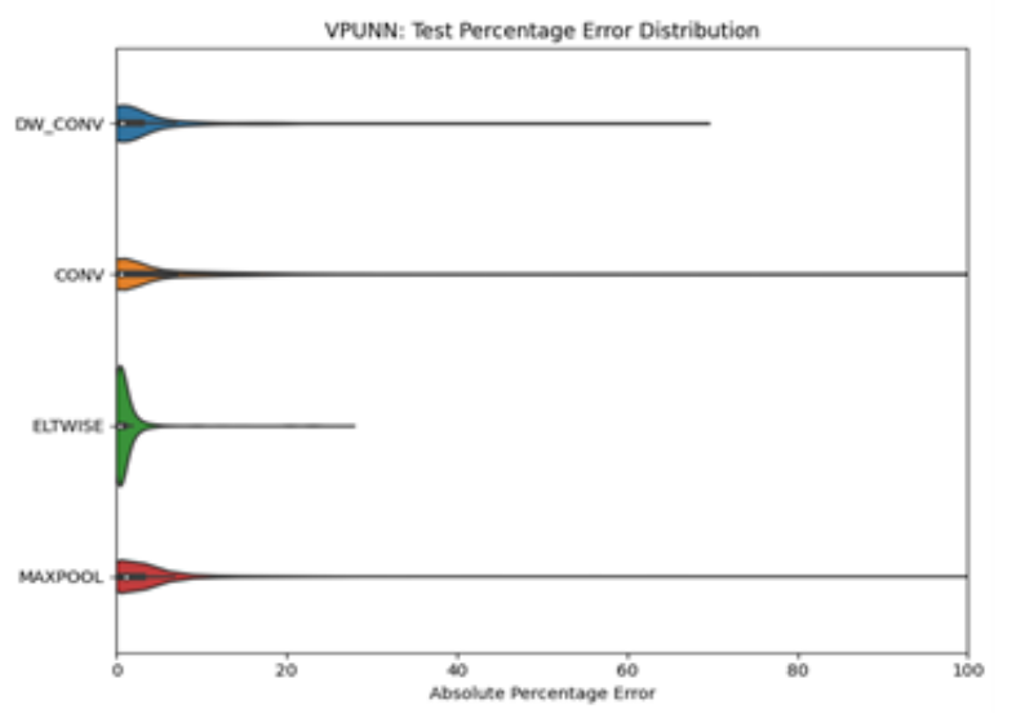}
    \caption{VPUNN absolute percentage Error}
    \label{fig.9}
\end{figure}

With the help of either hardware cost collection pipeline in section 3.3.1 or VPUNN in this section, we are able to generate a fairly accurate LUT for the HW-NAS model. We will show the experiment results using both methods in the next section.

\subsection{Results/Impact}
We in-cooperate ProxylessNAS with the LUT generation methods and search networks for Movidius VPU on various tasks including Image classification and super resolution. The searched networks are exported to ONNX format, then converted to OpenVINO IR, finally benchmarked on VPU. The experiment results show the advantage of the pipeline proposed in this whitepaper.

\subsubsection{ImageNet Classification}
We first apply the ProxylessNAS + Hardware cost collection pipeline and search for networks for ImageNet. In our experiments, the basic building block selected is inverted bottleneck residual block (MBConv) \cite{https://doi.org/10.48550/arxiv.2205.04586} from Mobilenet-v2.  This building block is well supported by VPU with good inference efficiency.  The super-net was built by a series of inverted bottleneck residual block with flexible depth, kernel size and expand ratio.  

\begin{figure}[h!]
    \centering
    \includegraphics[scale=1.2]{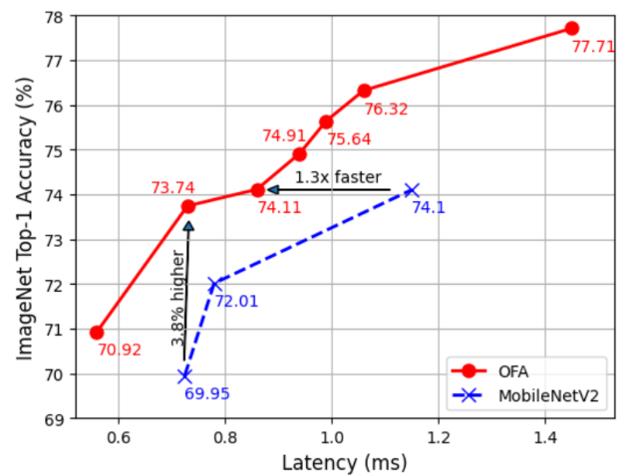}
    \caption{Network Latency and Top1 Accuracy}
    \label{fig.11}
\end{figure}

The results are tested on second generation Movidius VPU KeemBay. The accuracies of different models are as shown in fig.\ref{fig.11}. Compared with Mobilenet-v2 family on KeemBay, our methods can find network with 1162 fps @ 74.1 top1 accuracy at ImageNet which is 1.3x fps acceleration over Mobilenetv2-1.4 (870 fps @ 74.1) with close accuracy.   If we compare with Resnet structures, we are able to get higher acceleration 2.2 fps (684 fps@77.71 top1 vs Resnet50 320 @77 top1).

\begin{figure}[h!]
    \centering
    \includegraphics[scale=0.42]{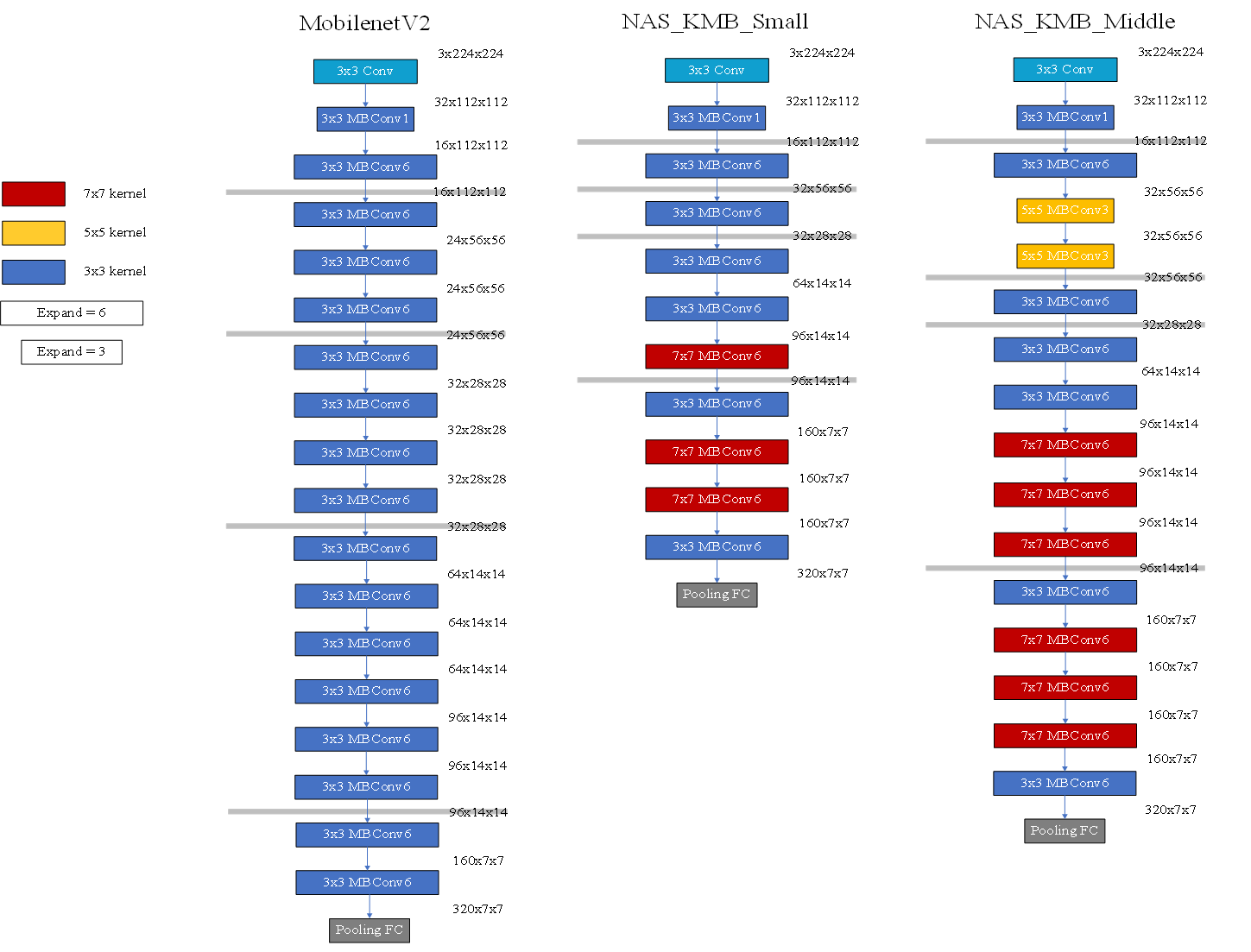}
    \caption{Topology of networks for ImageNet}
    \label{fig.12}
\end{figure}

\begin{figure}[h!]
    \centering
    \includegraphics[scale=1.1]{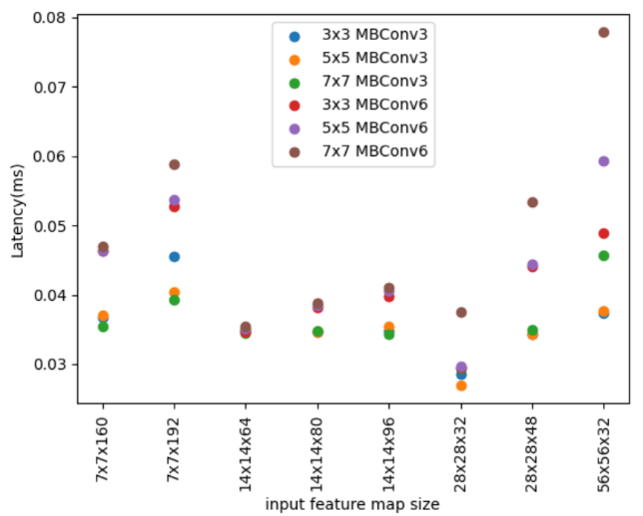}
    \caption{Block latency of different input feature map sizes}
    \label{fig.13}
\end{figure}

Fig.\ref{fig.12} shows the topology of Mobilenet-v2-1.0 and searched networks: NAS\_KMB\_Small and NAS\_KMB\_Middle. The accuracy of these models are shown in table.\ref{tb.1}. As shown, NAS\_KMB\_Small is shorter in length and NAS\_KMB\_Middle is longer in length. Both NAS searched networks have more large kernels in the latter stages of the network. This is because when featuremap is small, the benchmarked latency has very small difference for MBConv with different kernel sizes. But the large kernel size does have impact on accuracy. As show in fig.\ref{fig.13}, for MBConv6 with 7x7x160 input, deviation in latency is negligible between 3x3 vs 5x5 vs 7x7:  0.0463 vs 0.046416 vs 0.047016. 

% \usepackage{booktabs}
\begin{table}[h!]
    \centering
    \begin{tabular}{@{}llllll@{}}
    \toprule
    Network                   & Input & \#   of MBConv & Top1   @ ImageNet & Latency   on KMB (ms) & FPS on KMB \\ \midrule
    Moblinet-v2               & 224                & 19                              & 72.01             & 0.78                     & 1282            \\
    NAS\_KMB\_Small   (ours)  & 224                & 10                              & 70.92             & 0.56                     & 1785            \\
    NAS\_KMB\_Middle   (ours) & 224                & 15                              & 73.74             & 0.73                     & 1369            \\ \bottomrule
    \end{tabular}
    \caption{Detailed performance of classfication networks}
    \label{tb.1}
\end{table}

\subsubsection{Image super resolution}

We also apply the ProxylessNAS + VPUNN and search networks for super resolution tasks. As shown in fig.\ref{fig.13} left, original EDSR model has leaky ReLU as activation function and use Depth to Space to upsample the input to 2x resolution. However, these operators need to run on DSP on VPU, this will significantly slow down the inference speed. Thus in our NAS algorithm, apart from searching different kernel sizes for convolutions, we also search different upsampling methods (Nearest/Bi-linear/Depth to Space) and different activation functions (Leaky ReLU/ReLU). 

The results are tested on third generation Movidius VPU. The accuracy of the searched model and the original model are shown in table.\ref{tb.2}. We are able to achieve 6x speedup on next generation VPU with the proposed method.

\begin{figure}[h!]
    \centering
    \includegraphics[scale=0.5]{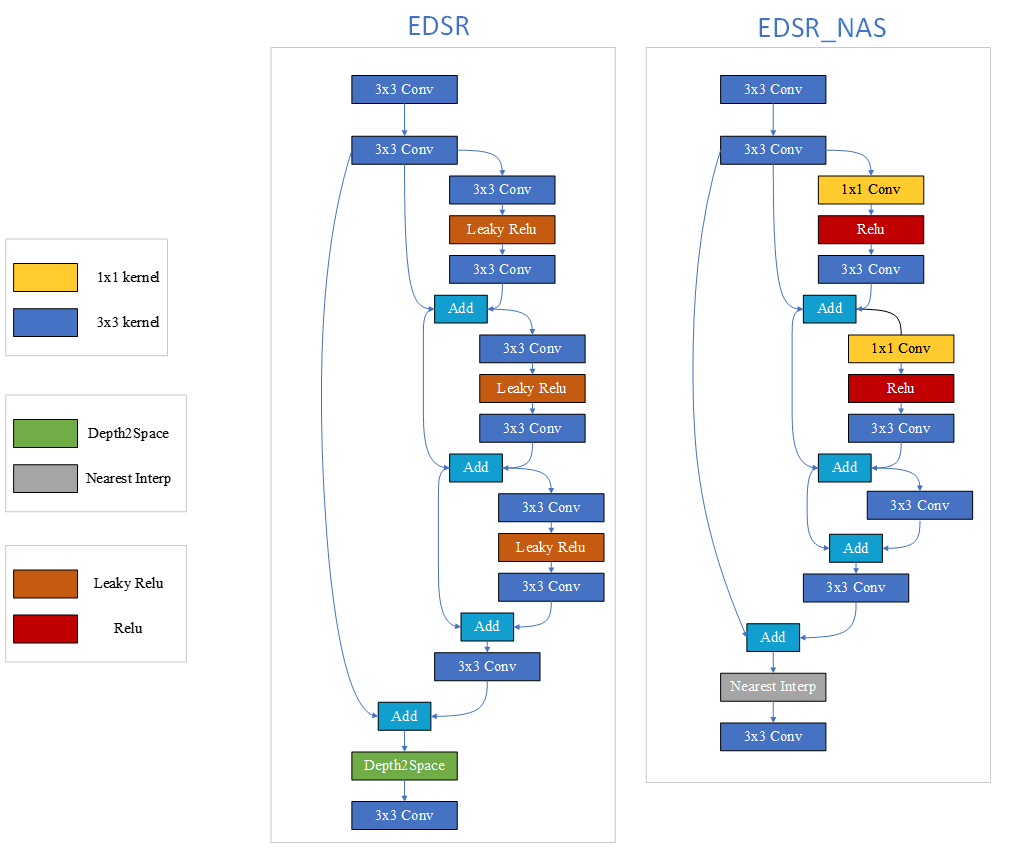}
    \caption{Topology of networks for super resolution}
    \label{fig.14}
\end{figure}

As shown in fig.\ref{fig.14}, the searched network use ReLU and Nearest Interpolation which are both very efficient on VPU hardware. The NAS networks also have more 1x1 kernels which is suppose to run faster than larger kernels especially when the feature map is huge.

% \usepackage{booktabs}
\begin{table}[h!]
    \centering
    \begin{tabular}{@{}lllll@{}}
    \toprule
    Network              & Input   Resolution & Upsample & PSNRY  & FPS speed up \\ \midrule
    EDSR3                & 360x640            & 2x       & 30.16  & NA   \\
    EDSR\_NAS\_S  (Ours) & 360x640            & 2x       & 32.647 & 6x           \\ \bottomrule
    \end{tabular}
    \caption{Detailed performance of super resolution networks}
    \label{tb.2}
\end{table}

In conclusion, we have the following intuitive insights for designing networks for VPU:
\begin{itemize}
    \item Avoid operations that will happen on DSP, like activations including Leaky Relu with Per-channel slope/ GeLU/ ... For Upsampling, avoid Interpolation with align corners/Depth to Space…
    \item Make the output channels of convolution 16x. In VPU, the minimal computation on channel dimension is 16, so non-16x channels will result in lower compute efficiency.
    \item Depthwise separate convolution vs Naïve convolution: 
        Depthwise plus Pointwise Convolution is faster in compute time and can achieve similar accuracy as normal convolution. But when activation need streaming, Depthwise Plust Pointwise might both bounded by DMA time. This will result in a longer latency than normal convolution.
\end{itemize}

\section{Further work}
The profiling method proposed in this white paper has limitation when it comes to networks with complex network topology. This is because we don't consider the impact of different scheduling methods. The latency model works with the assumption that the model is executed sequentially layer by layer and have no impact from the other layers, which is not true for cases like long residual connection and horizontal fusion of various operators. We are currently working on an efficient graph compiler  and back-end simulation tool that can potentially address this issue.

\section{Conclusion}
Designing hardware friendly networks can brings power and throughput benefit. Hardware-aware NAS technologies have been proven to be able to search models to meet accuracy and inference efficiency requirements on a given hardware.  This study illustrates how to use the ProxylessNAS and the two proposed hardware profile methods to design more efficient and accurate models on VPU.


\end{document}